\documentclass{article}

\PassOptionsToPackage{numbers, compress}{natbib}
\usepackage[preprint]{neurips_2026}

\usepackage[utf8]{inputenc}
\usepackage[T1]{fontenc}
\usepackage{hyperref}
\hypersetup{hidelinks}
\usepackage{url}
\usepackage{booktabs}
\usepackage{graphicx}
\usepackage{multirow}
\usepackage{amsfonts}
\usepackage{nicefrac}
\usepackage{microtype}
\usepackage[table]{xcolor}
\usepackage{array}
\usepackage{textcomp}
\usepackage{wrapfig}
\usepackage{needspace}
\usepackage{placeins}

\definecolor{deltaup}{RGB}{34,139,34}
\definecolor{deltadn}{RGB}{200,30,30}
\newcommand{\dplus}[1]{{\,\textcolor{deltaup}{\scriptsize+#1}}}
\newcommand{\dplusb}[1]{{\,\textcolor{deltaup}{\scriptsize\textbf{+#1}}}}
\newcommand{\dminus}[1]{{\,\textcolor{deltadn}{\scriptsize-#1}}}
\newcommand{\dzero}{{\,\textcolor{gray}{\scriptsize\textpm{}0.0}}}

\title{AdaCodec: A Predictive Visual Code \\ for Video MLLMs}

\author{%
{\small\textbf{Haowen Hou$^{1,2,3}$\thanks{Work done during an internship at JD.com.} \quad
Zhen Huang$^{2}$ \quad
Zheming Liang$^{2}$ \quad
Qingyi Si$^{3}$ \quad
Chenglin Li$^{2}$}} \\
{\small\textbf{Shuai Dong$^{2}$ \quad
Kele Shao$^{2}$ \quad
Ruilin Li$^{2}$ \quad
Dianyi Wang$^{2}$ \quad
Nan Duan$^{3}$ \quad
Jiaqi Wang$^{3,2}$\thanks{Corresponding author.}}} \\[0.8ex]
{\footnotesize\textnormal{$^{1}$Shanghai Jiao Tong University \quad
$^{2}$Shanghai Innovation Institute \quad
$^{3}$JD.com}} \\[0.35ex]
{\footnotesize\textnormal{\url{https://HaowenHou.github.io/AdaCodec-Page/}}}
}

\makeatletter
\patchcmd{\@maketitle}{\vskip 0.3in \@minus 0.1in}{\vskip 0.10in \@minus 0.05in}{}{}
\renewcommand{\@notice}{}
\makeatother

\newcommand{\splittag}[1]{{\footnotesize\ttfamily #1}}

\begin{document}

\maketitle

\vspace{-0.4em}
\noindent
\begin{minipage}{\textwidth}
{\centering
\includegraphics[width=0.98\textwidth]{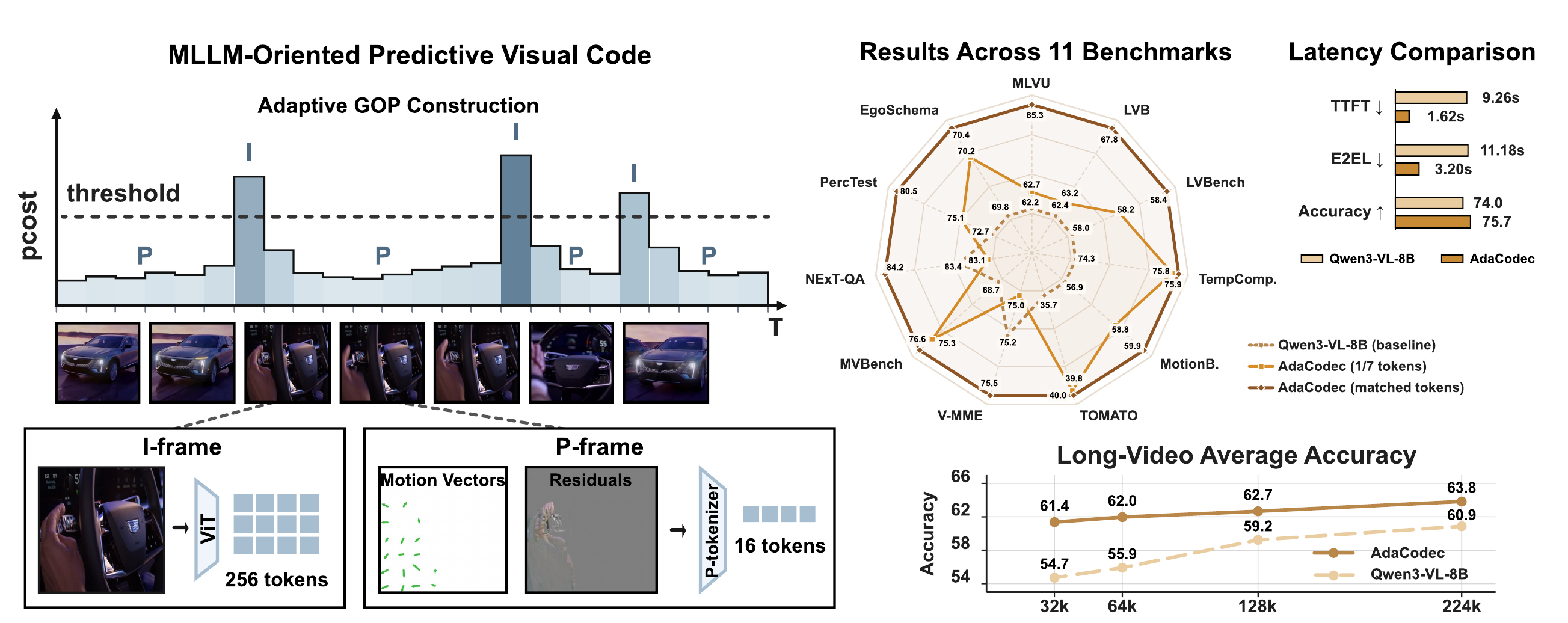}\par}
\vspace{3pt}
\refstepcounter{figure}\label{fig:teaser}
{\small
\leftskip=0pt
\rightskip=0pt
\parfillskip=0pt plus 1fil
\parindent=0pt
\noindent\textbf{Figure~\thefigure.} \textbf{AdaCodec treats the video MLLM visual interface as a predictive code:} a frame is encoded into full visual tokens only when it cannot be predicted from prior context, and intermediate frames are sent as compact motion-and-residual P-tokens. 
\textbf{Left:} AdaCodec splits the video into adaptive Groups of Pictures (GOPs), each containing one I-frame (intra-coded frame, encoded independently) followed by a chain of P-frames (predictive frames). AdaCodec places I-frames adaptively via a \emph{pcost} threshold on per-frame predictive cost, encodes each I-frame into full ViT tokens, and encodes each intermediate P-frame into fewer compact motion-and-residual tokens produced by the P-tokenizer. \textbf{Right:} AdaCodec matches or surpasses Qwen3-VL-8B on all eleven benchmarks even at $1/7$ the tokens, cuts time-to-first-token (TTFT) and end-to-end latency (E2EL) while raising average accuracy, and leads on long-video accuracy across token budgets from 32k to 224k.\par
}
\end{minipage}
\par
\vspace{0.5em}

\begin{abstract}
Video is temporally redundant: adjacent frames usually share most objects, background, and layout. Yet existing video multimodal large language models (video MLLMs) usually encode each sampled frame as an independent RGB image, causing visual tokens to repeat content already present in earlier frames. This suggests a more direct video interface: 
send a full reference frame only when the scene cannot be predicted well from prior context, and otherwise transmit a compact description of inter-frame changes. We call this interface a \emph{predictive visual code}, and instantiate it for video MLLMs as \textbf{AdaCodec}. AdaCodec spends full visual tokens on a reference frame only when its conditional predictive cost is high; otherwise, it encodes inter-frame changes, including motion and prediction residuals, as compact P-tokens. Across all eleven benchmarks, AdaCodec improves over the Qwen3-VL-8B per-frame RGB baseline at a matched visual-token budget. Even at $1/7$ the budget, AdaCodec with 32k tokens surpasses the 224k baseline on all long-video benchmarks; on five general-video benchmarks, it raises the average score while substantially cutting time-to-first-token from 9.26s to 1.62s.
\end{abstract}

\section{Introduction}
Video multimodal large language models (video MLLMs) are moving beyond short clips. They are increasingly used for workloads that require long temporal coverage, dense event tracking, and low response latency, including long-video understanding and temporal reasoning \citep{videochat,videochatgpt,videollama,videollava,chatunivi,llavaonevision,qwen2vl,videomme,mlvu}. Yet their visual interface remains largely unchanged: a video is sampled into RGB frames, each frame is encoded as an image, and the resulting visual tokens are concatenated into the language model context.

However, this per-frame interface is inefficient under temporal redundancy for video. Adjacent frames usually share most objects, background, and layout, so independent per-frame encoding repeatedly sends information that prior context already contains. Token cost then grows roughly linearly with the number of sampled frames. Under a finite context window, this redundancy creates a coverage--detail dilemma: sparse sampling misses short events and fine transitions, while dense sampling consumes context and increases latency.

Most efficiency methods reduce this pressure by selecting frames, compressing frame tokens, or managing frame-derived states through memory, long-context, and multi-rate designs \citep{keyvideollm,framevoyager,aks,mdp3,qframe,llamavid,longvu,dycoke,adaretake,framefusion,timechat,moviechat,malmm,videostreaming,longva,longvila,keyevl15}. These methods differ in where they save budget, but they share the same basic interface: retained visual evidence is still derived from independent RGB frames.
A related line has used codec-aware signals for efficient video understanding \citep{coviar,dmcnet,ema,copevideolm,remora}. These works show that codec structure can carry useful temporal evidence, but they keep the playback-oriented codec output fixed and learn modules that consume its extracted signals.
This leaves a sharper representation question: what visual representation should a video MLLM process, so the input removes temporal redundancy while preserving the evidence needed for reasoning?

We draw inspiration from predictive coding, where a system transmits errors from a prediction rather than the raw signal. This principle has biological grounding: the visual system is thought to encode prediction errors, the mismatch between expected and observed input, rather than the input itself \citep{rao_ballard_1999}. Modern video codecs use the same residual-coding idea in engineering: reference frames carry full content, while predictive frames carry motion and residual signals relative to a reference \citep{wiegand2003h264,sullivan2012hevc}. These systems have different objectives, but they share the same conditional structure: when nearby samples are redundant, the channel should carry what prediction fails to explain. Standard codecs, however, optimize for bitstreams and human-viewable reconstruction, not for visual tokens consumed by an LLM. We therefore redesign this mechanism as an MLLM interface for video understanding.

We present \textbf{AdaCodec}, a \textbf{predictive visual code for video MLLMs}. AdaCodec allocates full ViT tokens only to reference frames, and represents predictable intermediate frames with compact P-tokens derived from motion and residuals. Several design choices make the visual code MLLM-oriented, including a redesigned procedure for computing the predictive code, and a \emph{pcost}-driven reset that starts a new reference frame when prediction becomes costly. The MLLM therefore receives an interleaved stream of reference-frame tokens and compact P-tokens, instead of processing each sampled frame as a full RGB image. By eliminating redundant visual tokens before they enter the LLM, AdaCodec also substantially reduces inference latency.

Across eleven benchmarks, AdaCodec improves the performance-cost frontier. Under a matched 224k visual-token budget, it obtains the strongest open-source results in our comparison on all three long-video benchmarks and on two of the three temporal benchmarks. Under tighter budgets, AdaCodec at 32k visual tokens already surpasses the 224k Qwen3-VL-8B baseline on all long-video benchmarks. On the five general video-understanding benchmarks, AdaCodec uses 84.7\% fewer visual tokens, cuts time-to-first-token from 9.26s to 1.62s, and improves the average score. Our contributions are three-fold:

(1) We formulate \textbf{predictive visual code as the visual interface for video MLLMs}: a full reference frame is used only when conditional predictive cost is high, while predictable frames are compactly encoded through motion and residual cues.

(2) We build \textbf{AdaCodec}, including an MLLM-oriented predictive codec, a compact P-frame tokenizer, and a two-stage alignment pipeline that bridges the predictive code with existing MLLM architectures.

(3) Across \textbf{eleven benchmarks and controlled efficiency studies}, AdaCodec improves accuracy while substantially reducing both visual-token consumption and inference latency compared with per-frame RGB baselines. Ablations further validate each core design.
We will release the source code and model checkpoints.

\section{Related Work}

\subsection{Efficient Video Representations for Video MLLMs}
Video MLLMs encode each sampled frame into hundreds of visual tokens, so the visual sequence scales with video length and sampling density, therefore quickly dominating context length and compute. Prior work to alleviate this cost falls into three complementary directions.
A major line of work reduces cost through frame selection or frame-space subsampling \citep{keyvideollm,framevoyager,aks,mdp3,qframe}. Another line compresses tokens after frame encoding, including aggressive token compression, temporal pooling, dynamic token pruning or merging, and adaptive spatiotemporal compression \citep{llamavid,timechat,longvu,dycoke,adaretake,framefusion}. A third direction focuses on memory or long-context modeling, where sparse memory banks, online memory updates, and long-context adaptation improve scaling to longer videos \citep{moviechat,malmm,videostreaming,longva,longvila}. 
These methods are effective, but each retained frame is still encoded as an independent RGB image, leaving substantial redundancy among the retained frames.
AdaCodec instead replaces the per-frame interface itself, encoding predictable intervals as motions and residuals relative to a reference and spending a full reference frame only where prediction fails.

Two recent compression-oriented works provide useful context, although they do not target the same MLLM interface. OneVision-Encoder explores codec-aligned patch sparsity \emph{inside} the visual encoder via ``codec patchification'', targeting patch-level sparse computation and encoder efficiency \citep{ovencoder}. InfoTok learns adaptive discrete tokens for video reconstruction \citep{infotok}. AdaCodec is neither encoder-internal sparsity nor a generative reconstruction tokenizer; it is a predictive code at the visual interface that the MLLM consumes.

\subsection{Codec-based Video Representations}
Using compressed-domain signals for video understanding has a long history. Classical and modern codecs (e.g., AVC/H.264, HEVC/H.265, AV1) represent only sparse keyframes in full and model the remaining frames with predictive side information \citep{wiegand2003h264,sullivan2012hevc,chen2020av1}. Early studies used motion vectors as a low-cost surrogate for optical flow, and later methods such as CoViAR and DMC-Net modeled I-frame, motion, and residual modalities jointly for efficient action recognition \citep{zhang2016realtime,coviar,dmcnet}. Follow-up work improved multimodal fusion and compressed-domain transformer modeling \citep{teamnet,mmvit}. Beyond action recognition, compressed representations have been applied to object detection, video object segmentation, pose estimation, video question answering, and video captioning \citep{mmnet,xu2022compressedvos,fan2021pose,kim2021compressedvqa,shen2023compressedcaption}. These results support the view that motion and residual signals could encode useful localized temporal changes for higher-level reasoning.

Recently, codec-aware ideas have entered the video MLLM regime. EMA builds GOP-level representations from I-frames and motion vectors \citep{ema}. In concurrent work, CoPE-VideoLM uses standardized codec primitives from video streams and learns to align them with MLLM representations, while ReMoRa focuses on refining noisy block-motion representations and leveraging compressed-domain motion signals for long-video understanding \citep{copevideolm,remora}. These methods are complementary to AdaCodec but target a different design space: they treat a standards-compliant codec stream as fixed and learn how to consume it. AdaCodec instead redesigns the predictive code itself for MLLM consumption, so coding units, motion estimation, and reference-frame placement are all chosen for the downstream LLM rather than for human playback.

\section{Method}
\label{sec:method_full}
\begin{figure}[t]
\centering
\includegraphics[width=0.83\linewidth]{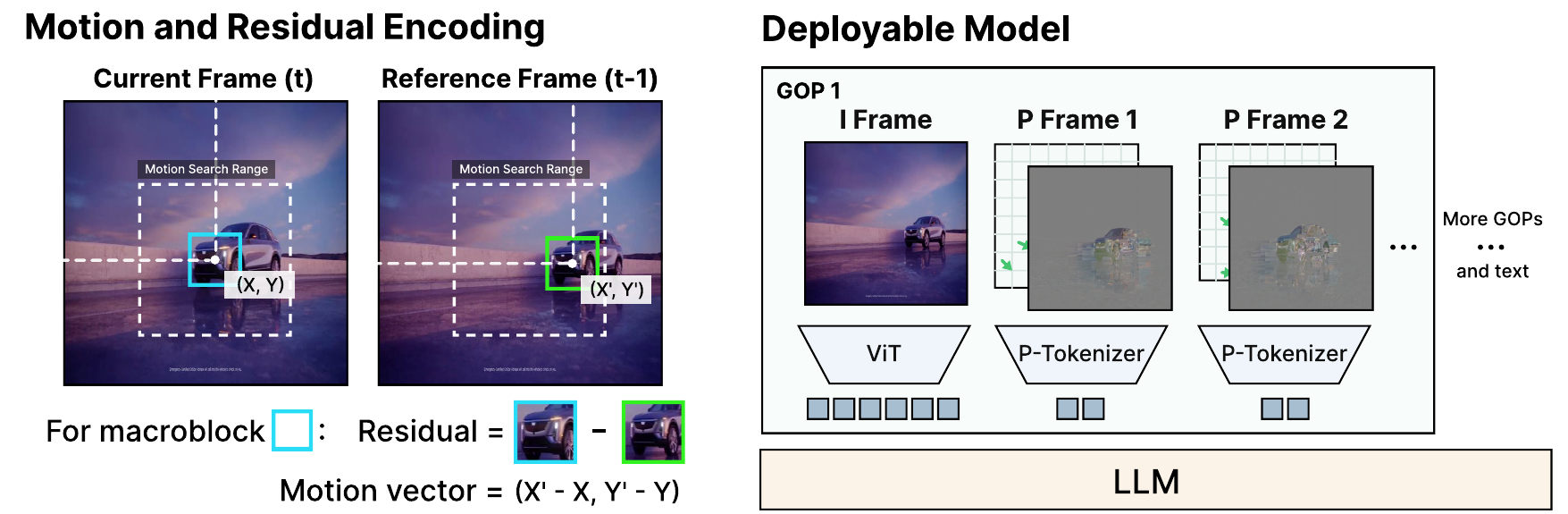}
\caption{\textbf{AdaCodec method overview.} \textbf{Left:} Motion-and-residual encoding for a P-frame. For each macroblock in the target frame, AdaCodec searches a local window in the reference frame for the best-matching block; the displacement gives the motion vector and the per-pixel difference gives the residual. \textbf{Right:} Deployable model. Each GOP encodes its I-frame with the ViT and each P-frame with the P-tokenizer.}
\label{fig:method_overview}
\end{figure}

Given a video $X=\{x_t\}_{t=0}^{T-1}$ with RGB frames $x_t\in\mathbb{R}^{H\times W\times 3}$, our goal is to construct a predictive visual code that video MLLMs can process effectively. AdaCodec is developed through three design aspects: (1) a predictive visual code that encodes intermediate frames as motion-and-residual updates, (2) a dual-branch visual-token pipeline for video MLLMs, and (3) two-stage training for P-frame representation learning and multimodal alignment of the visual code.
Figure~\ref{fig:method_overview} illustrates how AdaCodec encodes motion and residual for each P-frame and how the resulting tokens are consumed by the LLM.

\subsection{MLLM-Oriented Predictive Visual Code}

\paragraph{Codec preliminaries.}
A standard predictive codec stores occasional intra-coded keyframes (I-frames) in full and represents the remaining frames through motion-compensated prediction plus residual correction. The forward-predicted frames are P-frames, while bidirectionally predicted B-frames depend on both past and future references.

\begin{wraptable}{r}{0.55\linewidth}
\vspace{-2\baselineskip}
\centering
\scriptsize
\setlength{\tabcolsep}{3pt}
\renewcommand{\arraystretch}{1.}
\caption{Core redesigns from a playback-oriented codec to AdaCodec's MLLM-oriented predictive code.}
\label{tab:codec_redesign}
\begin{tabular}{@{}>{\raggedright\arraybackslash}m{0.14\linewidth}>{\raggedright\arraybackslash}m{0.34\linewidth}>{\raggedright\arraybackslash}m{0.46\linewidth}@{}}
\toprule
\textbf{Component} & \textbf{Playback-oriented codec} & \textbf{AdaCodec (MLLM-oriented)} \\
\midrule
Block partition & Heterogeneous block sizes chosen for bitrate. & Macroblocks aligned to the ViT patch grid, yielding more stable P-frame tokens. \\
Motion reference & Reference pictures selected under codec syntax. & Each P-frame is estimated from the immediately preceding sampled frame for larger temporal gaps. \\
Search window & Tuned to high-FPS playback. & Enlarged local window to absorb the larger displacement between low-FPS frames. \\
GOP scheduling & Separate content-analysis pass. & Reuses the predictive cost from motion search to trigger adaptive I-frame insertion. \\
\bottomrule
\end{tabular}
\vspace{-2.4\baselineskip}
\end{wraptable}

\paragraph{AdaCodec redesign for MLLM tokenization.}
AdaCodec adapts this predictive-coding paradigm to a video-MLLM interface. Standard codecs optimize a standards-compliant bitstream for transmission and human-perceived reconstruction, whereas a video MLLM consumes a visual-token sequence for reasoning.
Table~\ref{tab:codec_redesign} lists the core redesign choices that distinguish this code from a playback-oriented codec; the remaining redesign choices are in Appendix~\ref{app:codec_redesign}.

\paragraph{Motion and residual encoding.}
We partition $X$ into groups of pictures (GOPs), denoted by $\{\mathcal{G}_j\}_{j=1}^{J}$. For each GOP $\mathcal{G}_j=\{x_{s_j},\ldots,x_{e_j}\}$, the first frame $x_{s_j}$ is an I-frame, and each subsequent frame is a P-frame represented by motion vectors and residual signals relative to the preceding sampled frame. Thus, for every $t>s_j$, $x_t$ is predicted from $x_{t-1}$.

For a macroblock location $b \in \mathbb{Z}^2$ on the current frame, let $x_t^b$ be the current block and $x_{t-1}^{b+d}$ be the block at offset $d$ in the reference frame. AdaCodec searches in a local window $\mathcal{D}_b$ to find the best-matching block, as depicted in Figure~\ref{fig:method_overview}.
We define the candidate block prediction cost and select the motion vector by
\begin{equation}
\ell_t^b(d)=\mathrm{SAD}\!\left(x_t^b, x_{t-1}^{b+d}\right)+\lambda\|d\|_1,\qquad
m_t^b=\arg\min_{d\in\mathcal{D}_b}\ell_t^b(d).
\label{eq:motion_search}
\end{equation}
Here $d \in \mathbb{Z}^2$ is a 2D vector, $\mathrm{SAD}(\cdot,\cdot)$ is the sum of absolute pixel differences between two blocks, and the $\ell_1$ term mildly favors smaller displacements when several candidates match similarly well. The 2D motion vector $m_t^b$ points from the current 2D block location $b$ to the matched reference location $b+m_t^b$. The residual of this block is then the pixel difference between the current block and the reference block:
\begin{equation}
r_t^b = x_t^b - x_{t-1}^{b+m_t^b}.
\label{eq:residual_block}
\end{equation}
In practice, we use hexagonal search with local refinement to approximate the minimizer efficiently.

Each P-frame is represented by the signed residual $r_t\in\mathbb{R}^{H\times W\times 3}$ and block motion vectors $\{m_t^b\}_b$. Assigning each $m_t^b$ to every pixel in its block gives a 2-channel tensor $M_t\in\mathbb{R}^{H\times W\times 2}$. The P-frame input is the five-channel concatenation $u_t=[r_t,\,M_t]\in\mathbb{R}^{H\times W\times 5}$.

\paragraph{Adaptive GOP construction.}
The goal is to keep frames with low predictive cost in the same GOP and split when temporal prediction becomes less reliable.
Instead of fixed-length GOPs, we use a lightweight content-adaptive splitting rule. The same motion search above yields a frame-level predictive cost by summing the selected block costs:
\begin{equation}
\ell_t = \sum_b \ell_t^b(m_t^b).
\end{equation}
Here $\ell_t$ is the aggregate cost of predicting frame $x_t$ from $x_{t-1}$ under the selected block motions.
A large $\ell_t$ indicates that $x_t$ is poorly predicted as a P-frame and therefore contains substantial novel content, making it a strong candidate for a new I-frame.
This reuse lets AdaCodec make GOP decisions without a separate GOP-analysis pass, reducing duplicate computation and improving encoding speed.

We always designate frame $0$ as an I-frame. For each subsequent frame, we start a new GOP when $\ell_t>\gamma$, where $\gamma$ controls the GOP-length distribution and thus the token budget.
We choose $\gamma$ on the training split by targeting a median of $8$ P-frames per GOP, and reuse the resulting threshold for all training and evaluation runs.
To satisfy temporal-length constraints in MLLM training, we cap the number of P-frames per GOP by $K_{\max}$; once the cap is reached, we force an I-frame split.

\subsection{Dual-Branch Visual Tokenization Architecture}
The deployable AdaCodec architecture (Figure~\ref{fig:method_overview}) contains a reference-frame encoder $E_I$ and a P-frame tokenizer (P-tokenizer) $E_P$.
The P-tokenizer is initialized from a standard pretrained ViT. AdaCodec widens the pretrained ViT patch embedding from 3 to 5 input channels for $u_t=[r_t,\,M_t]$, copies the RGB kernels, and zero-initializes the two motion-vector channels. It then appends learnable tokens after the patch sequence and uses their output states as $E_P(u_t)$. This architecture adapts to any ViT-style visual encoder.
Details of the model architecture are described in Appendix~\ref{app:model_details}.

Given one GOP $\mathcal{G}=\{x_0,x_1,\ldots,x_K\}$, we treat I-frame $x_0$ as the reference frame, and then encode the reference frame and the $t$-th P-frame as
\begin{equation}
z^I = E_I(x_0)\in\mathbb{R}^{N_I\times d},\qquad
z_t^P = E_P(u_t) + e_t\in\mathbb{R}^{N_P\times d},\quad t=1,\ldots,K,
\end{equation}
where $e_t$ is the temporal position embedding and $N_P\ll N_I$. Therefore, a GOP with $K$ P-frames uses $N_I+KN_P$ tokens instead of $(K+1)N_I$, giving the token ratio analyzed in Section~\ref{sec:efficiency_latency} and Appendix~\ref{app:efficiency_details}.
The resulting reference and P-frame tokens are arranged into the LLM visual embedding space, then inserted into the multimodal prompt in temporal order.

\subsection{Two-Stage Training}
We train AdaCodec in two stages. Stage 1 learns the compact P-tokenizer under frozen visual supervision, whereas Stage 2 aligns the resulting visual code with the language model through multimodal training.

\paragraph{Stage 1: teacher-feature alignment for P-tokenizer.}
Each training sample contains one reference frame, $n$ intermediate P-frames, and one target frame. A frozen teacher visual encoder extracts
\begin{equation}
z^I = E_I(x_0),\qquad z^T = E_T(x_T),
\end{equation}
where $E_T$ shares weights with $E_I$. To train $E_P$, we attach an auxiliary feature predictor $H_\phi$, a lightweight transformer used only in Stage 1. Intermediate P-frames are encoded as $\{z_t^P\}_{t=1}^n$, and the auxiliary predictor outputs
\begin{equation}
\hat{z}^T = H_\phi\left(z^I,\{z_t^P\}_{t=1}^{n}\right).
\end{equation}
We optimize
\begin{equation}
\mathcal{L}_{\mathrm{stage1}} = \|\hat{z}^T-z^T\|_1 + \left(1-\cos(\hat{z}^T,z^T)\right).\label{eq:stage1_loss}
\end{equation}
During this stage, $E_I$ and $E_T$ remain frozen, while $E_P$ and $H_\phi$ are optimized.

\paragraph{Stage 2: multimodal alignment.}
After Stage 1, we keep the learned $E_P$ and discard the auxiliary predictor $H_\phi$. Under a fixed visual token budget, we uniformly sample multiple adaptive GOPs across the full video timeline and preserve their temporal order to form $\mathcal{V}_{\mathrm{code}}$. We then optimize the standard autoregressive next-token prediction loss:
\begin{equation}
\mathcal{L}_{\mathrm{stage2}} = -\sum_i \log p\left(y_i\mid y_{<i},\mathcal{V}_{\mathrm{code}},q\right),
\end{equation}
where $q$ is the text instruction and $y$ is the target response. We freeze all visual-side modules and update only the language model of the MLLM.

\section{Experiments}
\label{sec:experiments}
We evaluate AdaCodec on long-video, temporal, and general video-understanding benchmarks. The experiments proceed in three phases. We first present the main results: benchmark accuracy across eleven benchmarks (\S\ref{sec:main_results}) and system efficiency (\S\ref{sec:efficiency_latency}). Two analyses then inspect AdaCodec's behavior, examining how its accuracy advantage scales with the visual-token budget (\S\ref{sec:token_budgets}) and how its adaptive GOP construction tracks video content (\S\ref{sec:adaptive_gop_behavior}). Finally, we validate design necessity at two levels, the predictive coding (\S\ref{sec:predictive_vs_rate}) and our codec design choices (\S\ref{sec:codec_design_necessity}).

\subsection{Experimental Setup}
\label{sec:experimental_setup}
\paragraph{Model details and fairness protocol.}
We use Qwen3-VL-8B~\citep{qwen3vl} as the base MLLM. 
Because Qwen3-VL-8B ViT uses a temporal Conv3D visual stem, a spatial $2\times2$ merger, and DeepStack visual injection, AdaCodec must produce P-frame tokens that match these native interfaces; Appendix~\ref{app:model_details} describes this adaptation.
The details of the data sources and training hyperparameters are described in Appendix~\ref{app:training_details}.
For a fair comparison, we keep the number of visual tokens produced from each RGB frame the same for AdaCodec and Qwen3-VL-8B.
When the full frame sequence exceeds the visual context limit, for the baseline we follow the official uniform temporal sampling strategy, while for AdaCodec we uniformly sample GOPs over time.

\paragraph{Benchmarks.}
We organize benchmarks into three groups for presentation clarity.
{(1) Long-video}: MLVU~\splittag{test}~\citep{mlvu}, LongVideoBench~\splittag{val}~\citep{longvideobench}, LVBench~\splittag{test}~\citep{lvbench};
{(2) Temporal}: TempCompass~\splittag{test MCQ}~\citep{tempcompass}, MotionBench~\splittag{val}~\citep{motionbench}, TOMATO~\splittag{test}~\citep{tomato};
{(3) General video understanding}: Video-MME~\splittag{test}~\citep{videomme}, MVBench~\splittag{test}~\citep{mvbench}, NExT-QA~\splittag{test}~\citep{nextqa}, PerceptionTest~\splittag{val}~\citep{perceptiontest}, EgoSchema~\splittag{test}~\citep{egoschema}.
Benchmark descriptions are in Appendix~\ref{app:benchmark_details}.
We report macro-average accuracy on MLVU following standard practice. We use the \texttt{lmms-eval} toolkit for evaluation~\citep{zhang-etal-2025-lmms}. 

\subsection{Main Results Across Benchmarks}
\label{sec:main_results}
Table~\ref{tab:main_results} compares AdaCodec with the Qwen3-VL-8B baseline and other open-source models.
The baseline uses per-frame RGB input at 2\,FPS with a 224k visual-token budget. 
We report two operating points that answer complementary questions.

\textbf{(i) 1/7 token budget.}
This setting asks how performant AdaCodec is while using a significantly lower token budget saved by its predictive code.
Both methods consume the same frame sequence at 2\,FPS.
For long-video benchmarks, we explicitly cap AdaCodec at 32k visual tokens against the 224k-token RGB baseline; for temporal and general benchmarks, AdaCodec naturally uses about 1/7 of the baseline tokens through its compact representation.
This setting isolates token efficiency.

\textbf{(ii) Comparable token budget.} This setting asks whether the saved tokens can be converted into denser temporal evidence under a comparable total token budget.
For long-video benchmarks, we match the baseline's 224k visual-token budget.
For temporal and general benchmarks, we increase AdaCodec's frame rate from 2\,FPS to 16\,FPS, matching the baseline's total token use by statistics.
This setting isolates the gain from richer video coverage at fixed cost.

\begin{table*}[t]
\centering
\scriptsize
\setlength{\tabcolsep}{3pt}
\caption{Main benchmark results across long-video, temporal, and general video-understanding tasks. Higher is better. ``LVB'' denotes LongVideoBench, ``V-MME'' denotes Video-MME, and ``-'' indicates that the benchmark is not reported. \textbf{Bold} and \underline{underlined} numbers indicate the best and second-best results among open-source models, respectively. For external models, we use official reports when available; entries not reported there are taken from Molmo2~\citep{molmo2} or evaluated by us.}
\label{tab:main_results}
\resizebox{\textwidth}{!}{%
\begin{tabular}{l|ccc|ccc|ccccc}
\toprule
\multirow{2}{*}{Method} & \multicolumn{3}{c|}{\textbf{Long-video}} & \multicolumn{3}{c|}{\textbf{Temporal}} & \multicolumn{5}{c}{\textbf{General}} \\
& MLVU & LVB & LVBench & TempComp. & MotionB. & TOMATO & V-MME & MVBench & NExT-QA & PercTest & EgoSchema \\
\midrule
\rowcolor{gray!15}
\multicolumn{12}{l}{\textbf{Closed-source models}} \\
GPT-5~\citep{openai_gpt5} & 77.7 & 72.6 & 65.2 & 80.4 & 65.4 & 53.0 & 86.9 & 74.1 & 86.3 & 79.4 & 75.6 \\
GPT-5 mini~\citep{openai_gpt5} & 69.1 & 69.7 & 54.7 & 74.9 & 59.9 & 44.1 & 82.3 & 66.5 & 83.2 & 72.0 & 70.9 \\
Gemini 3 Pro~\citep{gemini3pro} & 75.7 & 75.9 & 77.0 & 82.8 & 62.6 & 48.3 & 87.5 & 70.4 & 84.3 & 77.6 & 68.9 \\
Gemini 2.5 Pro~\citep{gemini25} & 81.5 & 76.8 & 75.7 & 81.9 & 62.0 & 48.6 & 87.8 & 70.6 & 85.3 & 78.4 & 72.2 \\
Gemini 2.5 Flash~\citep{gemini25} & 75.1 & 73.1 & 64.9 & 80.2 & 59.3 & 39.1 & 84.2 & 67.0 & 81.8 & 74.7 & 70.2 \\
Claude Sonnet 4.5~\citep{claude_sonnet_45} & 64.0 & 65.1 & 50.5 & 72.8 & 58.5 & 39.6 & 80.5 & 62.1 & 79.2 & 64.3 & 73.1 \\
\midrule
\rowcolor{gray!15}
\multicolumn{12}{l}{\textbf{Open-source models}} \\
InternVL3.5-8B~\citep{internvl35} & 53.2 & 62.1 & 43.4 & 70.3 & 56.6 & 24.6 & 68.6 & 72.1 & 81.7 & 72.7 & 58.6 \\
Keye-VL-1.5-8B~\citep{keyevl15} & 53.8 & 66.0 & 42.8 & 75.5 & 55.1 & 33.0 & \textbf{76.2} & 56.9 & 75.8 & 64.2 & 56.3 \\
GLM-4.1V-9B~\citep{glm45v41v} & 56.6 & 65.7 & 44.0 & 72.3 & 59.0 & 30.0 & 75.6 & 68.4 & 81.3 & 74.2 & 62.6 \\
MiniCPM-V-4.5-8B~\citep{minicpmv45} & 60.6 & 63.9 & 50.4 & 72.7 & 59.7 & 29.8 & 73.5 & 60.5 & 78.8 & 70.9 & 49.6 \\
Eagle2.5-8B~\citep{eagle25} & 60.4 & 66.4 & 50.9 & 74.4 & 55.7 & 31.0 & 75.7 & 74.8 & 85.0 & 81.0 & \textbf{72.2} \\
PLM-8B~\citep{perceptionlm} & 52.6 & 56.9 & 44.5 & 72.7 & \underline{61.4} & 33.2 & 65.4 & \textbf{77.1} & 84.1 & \textbf{82.7} & 68.8 \\
LLaVA-Video-7B~\citep{llava_video} & 52.8 & 58.2 & 44.2 & 66.6 & 54.2 & 24.9 & 69.7 & 58.6 & 83.2 & 68.8 & 57.3 \\
VideoChat-Flash-7B~\citep{videochat_flash} & 56.0 & 64.7 & 48.2 & 69.4 & 60.6 & 32.5 & 69.7 & 74.0 & \underline{85.5} & 76.5 & 51.3 \\
Molmo2-8B~\citep{molmo2} & 60.2 & \underline{67.5} & 52.8 & 73.4 & \textbf{62.2} & 39.6 & \underline{75.8} & 75.9 & \textbf{86.2} & \underline{82.1} & 62.0 \\
Molmo2-O-7B~\citep{molmo2} & 55.2 & 63.7 & 49.6 & 73.0 & 60.6 & 36.2 & 69.2 & 74.8 & 84.3 & 79.6 & 56.8 \\
\midrule
\rowcolor{gray!15}
\multicolumn{12}{l}{\textbf{Codec-aware video MLLMs}} \\
CoPE-VideoLM-7B~\citep{copevideolm} & - & 56.9 & 46.4 & 68.9 & - & 28.3 & 69.4 & 61.9 & 82.1 & 70.3 & - \\
ReMoRa-7B~\citep{remora} & - & 60.8 & - & - & - & - & 64.4 & - & 84.2 & 67.7 & - \\
\midrule
\rowcolor{gray!15}
\multicolumn{12}{l}{\textbf{Ours (AdaCodec on Qwen3-VL-8B)}} \\
Qwen3-VL-8B~\citep{qwen3vl} & 62.2 & 62.4 & 58.0 & 74.3 & 56.9 & 35.7 & 75.2 & 68.7 & 83.4 & 72.7 & 69.8 \\
AdaCodec (1/7 token budget) & \underline{62.7}\dplus{0.5} & 63.2\dplus{0.8} & \underline{58.2}\dplus{0.2} & \underline{75.8}\dplusb{1.5} & 58.8\dplusb{1.9} & \underline{39.8}\dplusb{4.1} & 75.0\dminus{0.2} & 75.3\dplusb{6.6} & 83.1\dminus{0.3} & 75.1\dplusb{2.4} & 70.2\dplus{0.4} \\
AdaCodec (comparable token budget) & \textbf{65.3}\dplusb{3.1} & \textbf{67.8}\dplusb{5.4} & \textbf{58.4}\dplus{0.4} & \textbf{75.9}\dplusb{1.6} & 59.9\dplusb{3.0} & \textbf{40.0}\dplusb{4.3} & 75.5\dplus{0.3} & \underline{76.6}\dplusb{7.9} & 84.2\dplus{0.8} & 80.5\dplusb{7.8} & \underline{70.4}\dplus{0.6} \\
\bottomrule
\end{tabular}}
\end{table*}

\paragraph{Compactness preserves accuracy.}
At 1/7 of the baseline's tokens, AdaCodec maintains accuracy across all three categories. On long-video, results slightly exceed the baseline ($+0.5$, $+0.8$, $+0.2$ on MLVU, LongVideoBench, LVBench). The visual code thus preserves substantially more temporal information per token than per-frame RGB sampling. Temporal gains hold across TempCompass, MotionBench, and TOMATO ($+1.5$, $+1.9$, $+4.1$), so fine-grained motion is retained rather than traded for long-context coverage. On the five general benchmarks, AdaCodec gains on three (largest $+6.6$ on MVBench) and trails by at most 0.3 on the other two. Reduced visual tokens therefore do not come at the cost of general capability.

\paragraph{Extra coverage converts into accuracy.}
At a matched token budget, AdaCodec improves over the baseline on every benchmark, with the best open-source results on all three long-video benchmarks and on two of the three temporal benchmarks. On long-video, gains are $+3.1$, $+5.4$, $+0.4$ on MLVU, LongVideoBench, and LVBench. On temporal, gains are $+1.6$, $+3.0$, $+4.3$ across TempCompass, MotionBench, and TOMATO. The largest general-benchmark gains are $+7.9$ on MVBench and $+7.8$ on PerceptionTest. These gains under a matched token budget indicate that AdaCodec's predictive code converts its compactness into accuracy, rather than merely compressing the input.

\subsection{Efficiency and Latency}
\label{sec:efficiency_latency}
AdaCodec aims to improve both benchmark accuracy and system efficiency. We summarize the operating point below. The token-efficiency derivation, latency measurement protocol, codec-build overhead, and memory footprint are detailed in Appendix~\ref{app:efficiency_details}.

\Needspace{11\baselineskip}
\paragraph{Token efficiency.}
\begin{wraptable}[11]{r}{0.50\linewidth}
\vspace{-\baselineskip}
\centering
\small
\setlength{\belowcaptionskip}{7pt}
\caption{Token efficiency at the theoretical cap and over all evaluation videos. GOP length counts the I-frame and P-frames.}
\label{tab:token_stats}
\setlength{\tabcolsep}{0pt}
\renewcommand{\arraystretch}{1.12}
\begin{tabular*}{\linewidth}{@{\extracolsep{\fill}}lcc@{}}
\toprule
\raisebox{0.45\baselineskip}{Metric} & \shortstack{Theoretical\\cap} & \shortstack{All evaluation\\videos} \\
\midrule
GOP length$\uparrow$ & 17.00 & 10.21 \\
P-frames/GOP$\uparrow$ & 16.00 & 9.21 \\
Token ratio$\downarrow$ & 11.8\% & 15.4\% \\
Token reduction$\uparrow$ & 88.2\% & 84.6\% \\
\bottomrule
\end{tabular*}
\end{wraptable}
The maximum GOP length in AdaCodec is 17 frames (1 I-frame + 16 P-frames). Under this longest-GOP regime, AdaCodec incurs an $11.8\%$ token cost relative to per-frame RGB encoding. On real evaluation videos, content changes shorten some GOPs, giving an average GOP length of 10.21 frames (Table~\ref{tab:token_stats}). The measured token cost is still only $15.4\%$ of the baseline, an $84.6\%$ reduction. Section~\ref{sec:adaptive_gop_behavior} further analyzes this content-dependent GOP adaptation.

\Needspace{12\baselineskip}
\suppressfloats[t]
\paragraph{Latency and memory.}
\begin{wraptable}{r}{0.50\linewidth}
\vspace{-1.4\baselineskip}
\centering
\scriptsize
\setlength{\belowcaptionskip}{6pt}
\caption{Latency and peak-memory comparison. Score is the five-benchmark mean.}
\label{tab:latency}
\setlength{\tabcolsep}{1.2pt}
\renewcommand{\arraystretch}{1.03}
\begin{tabular*}{\linewidth}{@{\extracolsep{\fill}}lrrrrrr@{}}
\toprule
Method & Tokens$\downarrow$ & Build$\downarrow$ & TTFT$\downarrow$ & E2EL$\downarrow$ & Mem.$\downarrow$ & Score$\uparrow$ \\
\midrule
RGB base. & 55{,}893.2 & -- & 9.26s & 11.18s & \textbf{34.6~GB} & 74.0 \\
\textbf{AdaCodec} & \textbf{8{,}550.4} & 0.12s & \textbf{1.62s} & \textbf{3.20s} & 36.5~GB & \textbf{75.7} \\
\bottomrule
\end{tabular*}
\vspace{-1.4\baselineskip}
\end{wraptable}
Table~\ref{tab:latency} reports the latency and peak-memory comparison on the five general video-understanding benchmarks under matched hardware and decoding settings. Aggregated over 11{,}347 unique videos, AdaCodec uses 8{,}550.4 visual tokens per video on average against 55{,}893.2 for the per-frame RGB baseline ($84.7\%$ reduction), cuts time-to-first-token (TTFT) from 9.26s to 1.62s and total end-to-end latency (E2EL) from 11.18s to 3.20s, and raises the five-benchmark average score from 74.0 to 75.7. Codec-build denotes predictive code calculation and \emph{pcost}-based GOP splitting needed to construct the AdaCodec input on a consumer-level 16-core CPU; even when this 0.12s cost is charged to TTFT, AdaCodec remains $5.3\times$ faster than the baseline. The AdaCodec visual code thus delivers a Pareto improvement over per-frame RGB input: fewer visual tokens, faster response, and stronger downstream performance, at a one-time $+1.9$~GB peak-memory cost ($+5.5\%$).

\subsection{Performance across Visual Token Budgets}
\label{sec:token_budgets}
\begin{figure}[!ht]
\vspace{-0.2\baselineskip}
\centering
\makebox[\linewidth][c]{%
\begin{minipage}[b]{0.278\textwidth}
    \centering
    \includegraphics[width=\linewidth]{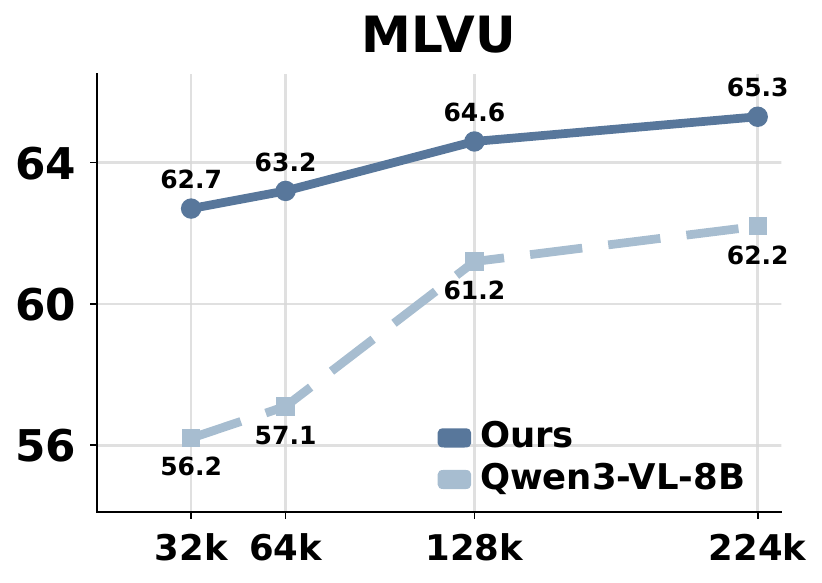}
\end{minipage}\hspace{0.012\textwidth}%
\begin{minipage}[b]{0.278\textwidth}
    \centering
    \includegraphics[width=\linewidth]{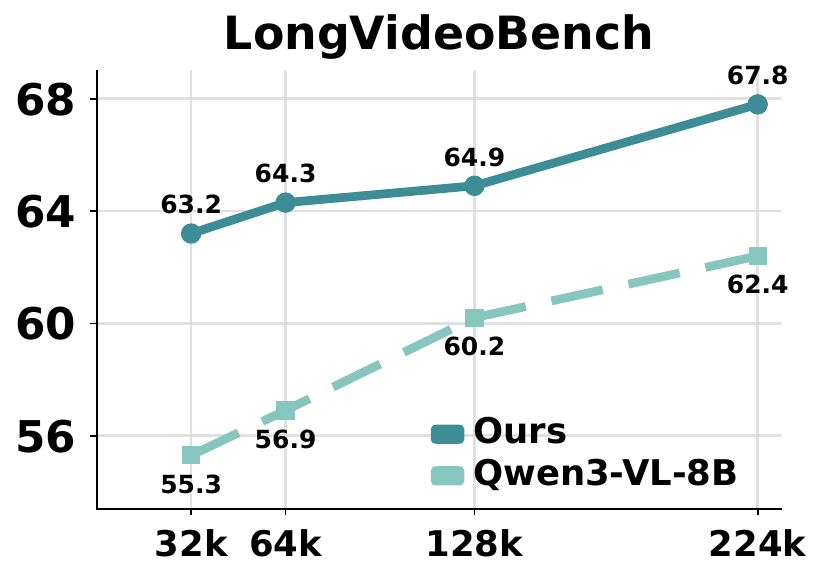}
\end{minipage}\hspace{0.012\textwidth}%
\begin{minipage}[b]{0.278\textwidth}
    \centering
    \includegraphics[width=\linewidth]{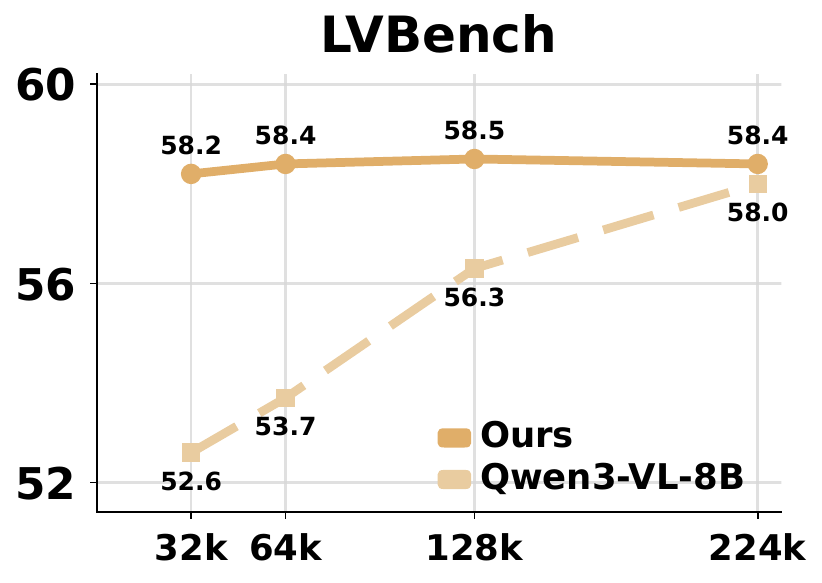}
\end{minipage}%
}
\caption{Long-video accuracy under visual-token budget sweeps.}
\label{fig:token_scaling}
\vspace{-0.4\baselineskip}
\end{figure}

On the three long-video benchmarks, we sweep visual-token budgets of 32k, 64k, 128k, and 224k. Temporal and general benchmarks are out of scope here because their videos consume fewer visual tokens overall. AdaCodec dominates the baseline across the full budget range on all three benchmarks (Figure~\ref{fig:token_scaling}), so the gain is not tied to a single operating point. At 32k visual tokens, AdaCodec already surpasses the 224k-token baseline, consistent with the claim that predictive coding makes more efficient use of the visual-token budget than independent RGB-frame encoding.

\subsection{Adaptive GOP Behavior}
\label{sec:adaptive_gop_behavior}
\begin{wrapfigure}{r}{0.46\textwidth}
\vspace{-0.8\baselineskip}
\centering
\includegraphics[width=\linewidth]{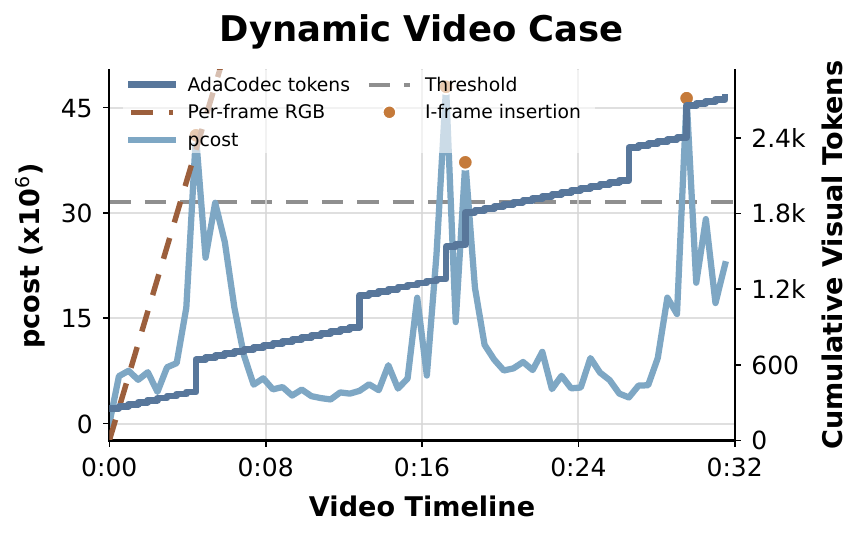}
\caption{Dynamic-case behavior under adaptive GOP construction. Spikes in \emph{pcost} trigger I-frame insertions, while P-frames keep AdaCodec's cumulative token growth far below per-frame RGB.}
\label{fig:dynamic_case_main}
\vspace{-0.9\baselineskip}
\end{wrapfigure}
Figure~\ref{fig:dynamic_case_main} first illustrates the adaptive I-frame reset mechanism on a dynamic example from NextQA. Several \emph{pcost} spikes cross the threshold, so AdaCodec inserts new I-frames before prediction errors accumulate. Between I-frame resets, intermediate frames remain compact P-token inputs; the cumulative token curve therefore grows much more slowly than per-frame RGB. The policy thus saves tokens in predictable intervals and spends full visual tokens when prediction becomes difficult.

The same rule yields different GOP patterns across video regimes. The global average GOP length of 10.21 from Table~\ref{tab:token_stats} hides a wide content-dependent spread. On MLVU~\splittag{test}~\citep{mlvu}, visually stable categories such as anomaly recognition (16.69) and tutorial-style videos (12.22) sustain long predictive chains, whereas egocentric (9.07) and dynamic content require more frequent I-frame refreshes. The anomaly-recognition gain comes from this behavior: AdaCodec preserves far more of the original sequence within the same context budget, reaching 71.8 against Qwen3-VL-8B's 51.2 ($+20.6$). Per-category breakdowns and additional case studies are in Appendix~\ref{app:adaptive_gop_details}.

\Needspace{14\baselineskip}
\suppressfloats[t]
\subsection{Necessity of Predictive Coding}
\label{sec:predictive_vs_rate}
\begin{wraptable}{r}{0.50\linewidth}
\vspace{-2.6\baselineskip}
\centering
\scriptsize
\setlength{\belowcaptionskip}{6pt}
\caption{Representation ablation under matched GOP coverage. All settings use full-RGB I-frames; only the P-frame input changes.}
\label{tab:interface_ablation}
\setlength{\tabcolsep}{1.5pt}
\renewcommand{\arraystretch}{1.04}
\begin{tabular}{@{}>{\raggedright\arraybackslash}p{0.26\linewidth}>{\raggedright\arraybackslash}p{0.26\linewidth}>{\centering\arraybackslash}p{0.105\linewidth}>{\centering\arraybackslash}p{0.16\linewidth}>{\centering\arraybackslash}p{0.14\linewidth}@{}}
\toprule
Setting & P-frame input & Long$\uparrow$ & Temporal$\uparrow$ & General$\uparrow$ \\
\midrule
I-only & Omitted & 52.7 & 48.5 & 68.2 \\
Per-frame RGB & Full RGB & 58.6 & 55.6 & 74.0 \\
Thumbnail P (untrained) & RGB thumbnail & 58.2 & 53.8 & 73.6 \\
Thumbnail P (trained) & RGB thumbnail & 61.4 & 54.4 & 73.2 \\
\textbf{AdaCodec} & \textbf{P-tokens} & \textbf{63.8} & \textbf{58.2} & \textbf{75.7} \\
\bottomrule
\end{tabular}
\vspace{-1.3\baselineskip}
\end{wraptable}
We test the necessity of predictive coding at the representation level. The GOP structure and full-RGB I-frame representation are fixed; only the P-frame representation changes (Table~\ref{tab:interface_ablation}).

\emph{Against I-only} ($+11.1$ / $+9.7$ / $+7.5$), P-tokens recover the motion and residual signal that a single keyframe discards. \emph{Against per-frame RGB} ($+5.2$ / $+2.6$ / $+1.7$), AdaCodec wins with a shorter visual prefix, so the gain is not from seeing more frames; the shorter prefix also reduces attention dilution from long token sequences. The all-RGB prefix can exceed Qwen3-VL-8B's native context window, so we use YaRN context extrapolation~\citep{peng2023yarn} for this setting.

\emph{Thumbnail P} provides a stricter control; we replace each P-frame with a low-resolution RGB thumbnail that Qwen3-VL-8B encodes into $N_P=16$ tokens, matching the AdaCodec per-frame budget.
AdaCodec exceeds the untrained Thumbnail P row by $+5.6$ / $+4.4$ / $+2.1$, and still exceeds the trained row by $+2.4$ / $+3.8$ / $+2.5$.
Low-resolution RGB loses fine visual detail and does not represent temporal changes explicitly. The remaining gap therefore comes from the predictive coding representation of AdaCodec, not only from adaptive token budget allocation.

\FloatBarrier
\subsection{Codec Design Ablations}
\label{sec:codec_design_necessity}
\Needspace{12\baselineskip}
\suppressfloats[t]
\begin{wraptable}{r}{0.50\linewidth}
\vspace{-2.6\baselineskip}
\centering
\scriptsize
\setlength{\belowcaptionskip}{6pt}
\caption{Core design ablations for AdaCodec. The default uses $16{\times}16$ macroblocks and adaptive GOP construction; deltas are relative to default.}
\label{tab:core_design_ablation}
\setlength{\tabcolsep}{1.5pt}
\renewcommand{\arraystretch}{1.04}
\begin{tabular}{@{}>{\raggedright\arraybackslash}p{0.35\linewidth}>{\centering\arraybackslash}p{0.18\linewidth}>{\centering\arraybackslash}p{0.19\linewidth}>{\centering\arraybackslash}p{0.17\linewidth}@{}}
\toprule
Setting & Long$\uparrow$ & Temporal$\uparrow$ & General$\uparrow$ \\
\midrule
\textbf{AdaCodec} & \textbf{60.3} & \textbf{56.2} & \textbf{74.1} \\
\midrule
Dynamic Macroblocks & 58.4\dminus{1.9} & 54.8\dminus{1.4} & 72.6\dminus{1.5} \\
\midrule
Fixed GOP, $n_P{=}8$ & 58.2\dminus{2.1} & 56.0\dminus{0.2} & 73.7\dminus{0.4} \\
Fixed GOP, $n_P{=}16$ & 59.7\dminus{0.6} & 55.4\dminus{0.8} & 72.3\dminus{1.8} \\
\bottomrule
\end{tabular}
\vspace{-1.3\baselineskip}
\end{wraptable}
Section~\ref{sec:predictive_vs_rate} tests the necessity of the predictive coding interface itself; we now test the necessity of two specific codec-design choices: ViT-aligned macroblocks and \emph{pcost}-guided GOP construction. Results are reported in Table~\ref{tab:core_design_ablation}; the full ablations are in Appendix~\ref{app:ablation_details}.

Replacing ViT-aligned $16{\times}16$ macroblocks with H.264-style dynamic partitions lowers all three category averages by $1.4$--$1.9$ points, indicating that the MLLM benefits from motion fields aligned to the ViT patch grid. Fixed GOP schedules also trail the adaptive \emph{pcost} policy on every category, with the largest gap of $2.1$ on long-video against $n_{P}{=}8$. The remaining axes, $N_P$, $K_{\max}$, and the threshold target, stay within $\pm 1.1$ around the default in Appendix~\ref{app:ablation_details}, so the main sensitivity comes from the MLLM-oriented codec design rather than from narrow hyperparameter tuning.

\section{Conclusion}
\paragraph{Limitations.}
Our experiments use a fixed input resolution; we leave dynamic resolution input to future work. AdaCodec also uses a uniform per-P-frame token budget ($N_P=16$), and adapting it to per-frame motion or residual complexity could further improve the efficiency--accuracy frontier. Finally, we do not evaluate AdaCodec on streaming video, although its causal I/P structure with incremental motion search has the potential in principle to support streaming with high frame rates and substantially reduce latency relative to per-frame RGB baselines.

\paragraph{Conclusion.}
We introduced AdaCodec, an MLLM-oriented redesign of predictive visual code. Rather than adapting a fixed playback codec stream, AdaCodec provides a series of codec redesigns for the visual-token interface of video MLLMs. It allocates full ViT tokens to high-cost reference frames and represents predictable intermediate frames with compact motion-and-residual P-tokens. This design removes repeated visual evidence before it enters the LLM context while preserving temporal changes needed for reasoning. Across long-video, temporal, and general video-understanding benchmarks, AdaCodec consistently improves over a per-frame RGB interface under matched or smaller visual-token budgets, with substantially lower response latency. Ablations show that both predictive coding and the MLLM-oriented codec redesign are necessary for these gains.

{\small
\bibliographystyle{unsrtnat}
\bibliography{references}
}

\clearpage
\appendix
\section*{Appendix}

\section{Codec Redesign Details}
\label{app:codec_redesign}

Table~\ref{tab:codec_redesign_full} reports every component in which AdaCodec departs from a standards-compliant playback codec. The first four rows reproduce the core redesigns highlighted in the main text (Section~\ref{sec:method_full}); the last three (color space, frame types, entropy coding) are the choices that follow from targeting an MLLM-token sequence rather than a playback bitstream.

\begin{table}[h]
\centering
\small
\setlength{\tabcolsep}{4pt}
\renewcommand{\arraystretch}{1.15}
\caption{Full component-wise comparison between a playback-oriented codec and AdaCodec. The upper block reproduces the core redesigns of Section~\ref{sec:method_full}; the lower block lists the additional configuration choices specific to the MLLM-token target.}
\label{tab:codec_redesign_full}
\begin{tabular}{>{\raggedright\arraybackslash}p{0.18\linewidth}>{\raggedright\arraybackslash}p{0.38\linewidth}>{\raggedright\arraybackslash}p{0.40\linewidth}}
\toprule
\textbf{Component} & \textbf{Playback-oriented codec} & \textbf{AdaCodec (MLLM-oriented)} \\
\midrule
Block partition & Heterogeneous block sizes chosen for bitrate. & Macroblocks aligned to the ViT patch grid, yielding more stable P-frame tokens. \\
Motion reference & Reference pictures selected under codec syntax. & Each P-frame is estimated from the immediately preceding sampled frame to handle high motions in larger temporal gaps. \\
Search window & Tuned to high-FPS playback. & Enlarged local window to absorb the larger displacement between low-FPS frames. \\
GOP scheduling & Separate content-analysis pass. & Reuses the predictive cost from motion search to trigger adaptive I-frame insertion for efficiency. \\
\midrule
Color space & YCbCr with chroma subsampling. & RGB, matching vision-backbone inputs. \\
Frame types & I, P, and bidirectional B. & I/P only; each predictive frame uses past context, compatible with streaming. \\
Entropy coding & Required for bitstreams. & Omitted; the output is a token sequence, not a bitstream. \\
\bottomrule
\end{tabular}
\end{table}

\section{Model Implementation Details}
\label{app:model_details}

\paragraph{P-token construction.}
For each P-frame, AdaCodec first forms the five-channel tensor $u_t=[r_t,\,M_t]\in\mathbb{R}^{H\times W\times 5}$ from the residual and motion vectors defined in Section~\ref{sec:method_full}.
The P-tokenizer uses an architecture-matched ViT backbone initialized from a pretrained visual encoder, with the patch-embedding input widened from three to five channels.
The RGB kernels are copied from the pretrained stem, and the two motion-vector channels are initialized to zero before training.
A standard ViT admits extra learnable tokens appended after the patch sequence without any change to the backbone, so we attach $N_L$ learnable latent tokens to the patch tokens before the visual transformer:
\begin{equation}
\tilde{z}_t^P = F_P\!\left([B_P(u_t), q_1,\ldots,q_{N_L}]\right)_{q_1:q_{N_L}},
\end{equation}
where $B_P$ is the patch embedding, $F_P$ is the residual visual backbone, and the subscript selects the output states of the appended latent tokens.
A block attention mask prevents patch tokens from attending to the latent tokens, while the latent tokens attend to all patch tokens.
This mask preserves the pretrained patch-token computation while letting the latent tokens aggregate information from the full predictive representation.
The resulting P-tokens are learned summary tokens conditioned on the residual-and-motion representation, rather than sampled image patches.

\paragraph{Stage 1 reconstruction module.}
Stage 1 trains the P-frame tokenizer $E_P$ through an auxiliary feature predictor $H_\phi$ that maps the I-frame embedding and the P-token states back to the teacher feature at the target frame (Section~\ref{sec:method_full}).
The reconstruction head is active only in Stage 1 and is removed before Stage 2, leaving only $E_P$ for downstream multimodal alignment.

\paragraph{Qwen3-VL ViT interface.}
Qwen3-VL introduces three changes to the standard ViT visual encoder that the P-tokenizer must match:
(a) the patch embedding is a Conv3D with temporal size two, which lets the stem take two frames jointly rather than a single image;
(b) a $2\times2$ spatial merger at the output reduces every four adjacent tokens into one merged visual token;
and (c) three intermediate layers are exported through a DeepStack visual-injection path into the first three language-model layers.
AdaCodec adapts to each change without modifying the pretrained backbone.
For (a), we encode a single I-frame or P-frame tensor by duplicating it along the temporal dimension and feeding the pair to the original Conv3D stem.
For (b), we apply the same $2\times2$ merger to both streams: at $512\times512$ inputs with patch size $16$, an I-frame yields $32\times32$ patch tokens and $N_I=256$ merged visual tokens, and the $N_L$ latent P-token states are arranged as a square grid and passed through the same merger, giving $N_P=N_L/4$ merged P-tokens per P-frame.
For (c), the P-tokenizer exposes the matching intermediate layers alongside the final output, and the Qwen3-VL forward pass feeds them through the native DeepStack injection path.

\section{Training Details}
\label{app:training_details}

Both training stages use the same public video-instruction data source.
Stage~1 samples tuples containing a reference frame, a sequence of intermediate P-frames, and a target frame from the training videos for teacher-feature alignment; Stage~2 uses the paired instruction-response examples for next-token training.
The Stage~2 instruction-tuning mixture contains 3{,}904{,}313 training examples.
Table~\ref{tab:training_sources} lists the shared public source families.
When a dataset has official train/validation/test partitions, both stages use only the training partition.

\begin{table}[t]
\centering
\small
\setlength{\tabcolsep}{4pt}
\caption{Shared public source families for Stage~1 and Stage~2 training. Molmo2 contributes its AskModelAnything, VideoCapQA, and VideoSubtitleQA subsets. Split handling is described in the text.}
\label{tab:training_sources}
\begin{tabular}{>{\raggedright\arraybackslash}p{0.30\linewidth}>{\raggedright\arraybackslash}p{0.30\linewidth}>{\raggedright\arraybackslash}p{0.30\linewidth}}
\toprule
\multicolumn{3}{c}{Dataset} \\
\midrule
ActivityNet-QA~\citep{activitynetqa} & LongVILA~\citep{longvila}            & TextVR~\citep{textvr} \\
Charades~\citep{charades}            & LSMDC~\citep{lsmdc}                  & TGIF-QA~\citep{tgifqa} \\
CharadesEgo~\citep{charadesego}      & Mementos~\citep{mementos}            & TVQA~\citep{tvqa} \\
CLEVRER~\citep{clevrer}              & Molmo2~\citep{molmo2}                & Video-ChatGPT~\citep{videochatgpt} \\
DiDeMo~\citep{didemo}                & NExT-QA~\citep{nextqa}               & Video-R1~\citep{video_r1} \\
Ego4D~\citep{ego4d}                  & OneThinker~\citep{onethinker}        & VideoChat2~\citep{mvbench} \\
EgoClip~\citep{egovlp}               & Perception Test~\citep{perceptiontest} & VideoEspresso~\citep{videoespresso} \\
EgoProceL~\citep{egoprocel}          & QuerYD~\citep{queryd}                & VideoGPT+~\citep{videogptplus} \\
EgoQA~\citep{egoqa}                  & ShareGPT4Video~\citep{sharegpt4video} & Vript~\citep{vript} \\
HiREST~\citep{hirest}                & SQA3D~\citep{sqa3d}                  & VTG-IT~\citep{vtgit} \\
LLaVA-Video-178K~\citep{llava_video} & STAR~\citep{star}                    & YouCook2~\citep{youcook2} \\
\bottomrule
\end{tabular}
\end{table}

\paragraph{Training hyperparameters.}
Table~\ref{tab:training_hyperparams} reports the training hyperparameters for the two training stages.
During stage~2 training, we train with 64k visual token budget for 40{,}000 steps, and then train on 224k for 5{,}000 steps.

\begin{table}[h]
\centering
\small
\setlength{\tabcolsep}{4pt}
\renewcommand{\arraystretch}{1.12}
\caption{Training hyperparameters for AdaCodec.}
\label{tab:training_hyperparams}
\begin{tabular}{>{\raggedright\arraybackslash}p{0.22\linewidth}>{\raggedright\arraybackslash}p{0.35\linewidth}>{\raggedright\arraybackslash}p{0.35\linewidth}}
\toprule
Setting & Stage~1 & Stage~2 \\
\midrule
Optimizer & AdamW, $\beta=(0.9,0.95)$, $\epsilon=10^{-8}$, weight decay 0.01 & AdamW, $\beta=(0.9,0.999)$, $\epsilon=10^{-8}$, weight decay 0 \\
Learning rate & $1{\times}10^{-4}$ peak; cosine decay to $1{\times}10^{-5}$; 4{,}000 warmup steps & $1{\times}10^{-5}$ peak; cosine decay; 0.1 warmup ratio \\
Global batch size & 128 & 128 \\
Optimization steps & 130{,}000 & 45{,}000 \\
\bottomrule
\end{tabular}
\end{table}

\paragraph{Compute resources.}
The training runs on 64 NVIDIA H800 GPUs and span approximately 12 days of wall-clock time. Latency measurements in Section~\ref{sec:efficiency_latency} use a single H800 for prefill and decoding, and the codec-build step is timed on a 16-core consumer CPU. The full research effort uses more resources, including preliminary runs and ablation experiments.

\section{Benchmark Descriptions}
\label{app:benchmark_details}

We evaluate on eleven public video benchmarks described below.

\paragraph{Long-video benchmarks.}
MLVU samples long-form videos from heterogeneous genres such as movies, surveillance, egocentric clips, cartoons, and gameplay, and pairs each clip with multi-task evaluation across varying durations and task types~\citep{mlvu}.
LongVideoBench is a multiple-choice QA benchmark for video-language interleaved inputs up to one hour long; its referring-reasoning questions ask the model to retrieve the relevant video context and reason over detailed multimodal evidence~\citep{longvideobench}.
LVBench targets extreme long-video understanding on public videos spanning hours, with tasks designed around long-term memory, extended comprehension, and information extraction~\citep{lvbench}.

\paragraph{Temporal benchmarks.}
TempCompass evaluates temporal perception over attributes such as speed and direction, and requests answers in multiple formats. To force genuine temporal reasoning, it pairs videos that hold their static content fixed while diverging in time-varying attributes, so single-frame cues and language priors cannot suffice~\citep{tempcompass}.
MotionBench assesses how well video models comprehend fine-grained motion. The benchmark draws clips from heterogeneous sources and partitions evaluation into six motion-oriented question categories, each targeting a specific aspect of motion-level perception~\citep{motionbench}.
TOMATO targets visual temporal reasoning. It defines six task types: action count, direction, rotation, shape and trend, velocity and frequency, and visual cues. TOMATO is designed so that the answer requires more than a single frame, the original frame order, and evidence drawn from across the clip~\citep{tomato}.

\paragraph{General video-understanding benchmarks.}
Video-MME evaluates MLLMs on short, medium, and long videos drawn from six visual domains and 30 subfields, testing whether models can answer video-centered questions across diverse content and temporal scales~\citep{videomme}.
MVBench converts public video annotations into 20 multiple-choice tasks that require temporal understanding beyond a single frame~\citep{mvbench}.
NExT-QA defines three video-QA question types (causal action reasoning, temporal action reasoning, and common-scene comprehension), framed to move beyond surface scene description toward explanation of actions~\citep{nextqa}.
PerceptionTest probes the transfer of pre-trained video models. The benchmark pairs four skill areas (memory, abstraction, physics, and semantics) with four reasoning types (descriptive, explanatory, predictive, and counterfactual), administered jointly over video, audio, and text inputs~\citep{perceptiontest}.
EgoSchema is built from Ego4D three-minute egocentric clips and asks five-option questions whose evidence spans long temporal certificate sets, making it a test of first-person video reasoning~\citep{egoschema}.

\section{Efficiency and Latency Details}
\label{app:efficiency_details}

\paragraph{Token-efficiency derivation.}
For a GOP with one I-frame and $K$ P-frames, the visual-token ratio against a per-frame RGB input is
\begin{equation}
\rho(K)=\frac{N_I+K N_P}{(K+1)N_I},
\label{eq:token_ratio}
\end{equation}
where $N_I$ and $N_P$ are tokens per I-frame and per P-frame. With AdaCodec's default setup, $N_I=256$ and $N_P=16$. Under the maximum predictive chain length used in our implementation (16 P-frames per GOP), the architectural minimum ratio is $\rho(16)=0.118$, an $88.2\%$ reduction. In practice, GOP length is content-dependent. Aggregated over all evaluation videos, the realized average GOP length is 10.21 frames, i.e., 9.21 P-frames per GOP. Substituting the empirical average into Eq.~(\ref{eq:token_ratio}) yields a $15.4\%$ visual-token ratio, an $84.6\%$ reduction relative to per-frame RGB input.

\paragraph{Latency measurement protocol.}
We evaluate runtime on the five general video-understanding benchmarks listed in Section~\ref{sec:experimental_setup} using time-to-first-token (TTFT) and total end-to-end latency (E2EL). All methods run on the same hardware with batch size 1, the same prompt template, identical decoding hyperparameters, the same 64 generated tokens, and the same input resolution. AdaCodec uses its I/P-frame visual code, whereas the per-frame RGB baseline feeds every frame as an RGB image. Aggregated over 11{,}347 unique videos, AdaCodec uses 8{,}550.4 visual tokens per video on average, whereas the per-frame RGB baseline uses 55{,}893.2 ($84.7\%$ reduction). Reducing visual prefilling lowers TTFT by design; AdaCodec also reduces total generation latency while improving downstream score, so the gain cannot be explained by discarding visual information for speed.

\paragraph{Codec-build overhead.}
Constructing the AdaCodec visual code online incurs a one-time codec-build step per video: predictive coding calculation and \emph{pcost}-driven GOP splitting. On a 16-core consumer CPU, codec-build takes 0.12s per video, about $7\%$ of AdaCodec's 1.62s TTFT. Folding it back into TTFT yields 1.74s, still $5.3\times$ faster than the per-frame RGB baseline's 9.26s, and E2EL retains a $3.4\times$ gap. The overhead is small in absolute terms and does not change the latency advantage of the AdaCodec visual code.

\paragraph{Memory footprint.}
AdaCodec adds one ViT-sized visual branch for the P-frame tokenizer (about 576M parameters, $7\%$ of the 8.14B-parameter backbone). Measured under FP16, AdaCodec increases peak GPU memory over the per-frame RGB baseline by 1.9~GB ($5.4\%$).

\section{Adaptive GOP Behavior Details}
\label{app:adaptive_gop_details}

\begin{figure}[h]
\centering
\includegraphics[width=\linewidth]{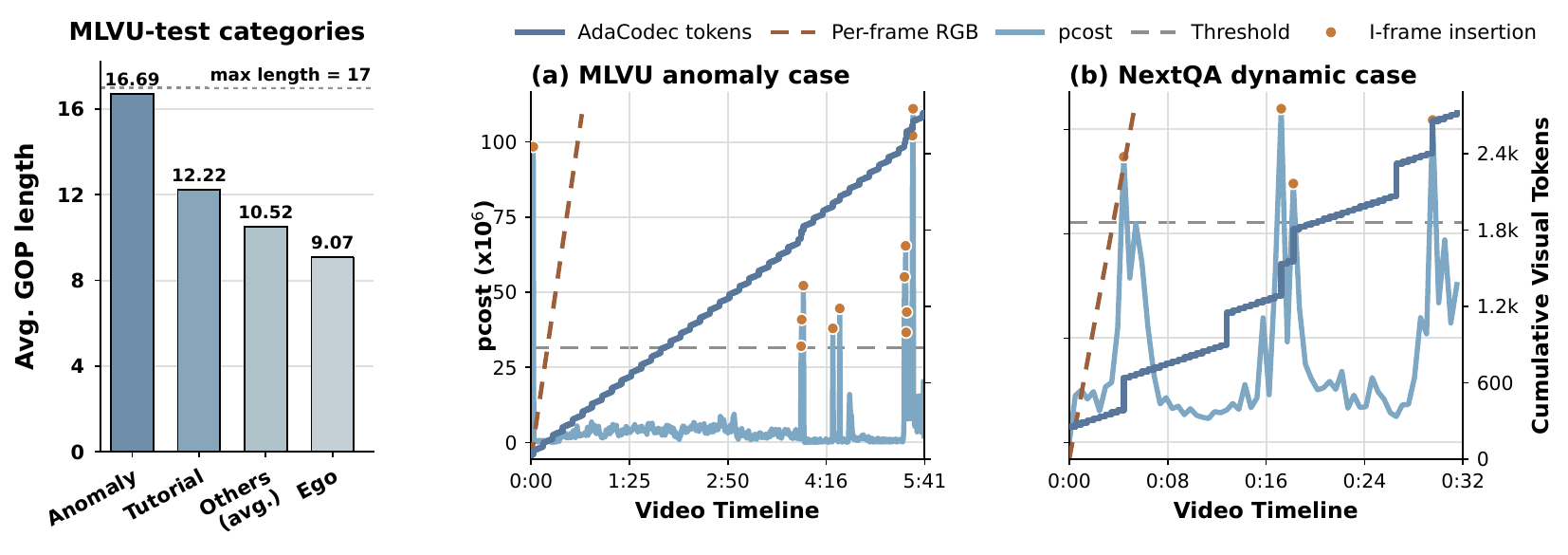}
\caption{Adaptive GOP behavior in AdaCodec. Left: average GOP length for representative MLVU-test categories; ``Others'' averages the remaining official categories. Middle: an MLVU anomaly case. Right: a dynamic case from NextQA. The per-frame RGB trajectory uses 256 tokens per frame and quickly exits the plotting range, highlighting how AdaCodec avoids runaway token growth.}
\label{fig:adaptive_behavior}
\end{figure}

\paragraph{Per-category mechanism.}
We take MLVU~\splittag{test}~\citep{mlvu} as a case study. The left panel of Figure~\ref{fig:adaptive_behavior} reports the three representative categories discussed in Section~\ref{sec:adaptive_gop_behavior} (anomaly recognition, tutorial-style videos, and ego reasoning) and averages the remaining official categories into ``Others.'' The category-level variation is consistent with the composition of MLVU. Anomaly recognition is built from surveillance-style videos with relatively fixed cameras and slowly evolving scenes. Tutorial videos often exhibit a more stable camera setup and step-wise temporal structure, which also permits longer GOPs. Ego reasoning is derived from egocentric first-person videos with frequent viewpoint changes and stronger camera motion. The averaged ``Others'' category lies between these cases, which is expected because it mixes videos with different levels of camera motion and event density. Thus, the per-category trend provides an interpretable explanation for the adaptive behavior: it allocates more visual tokens to temporally unstable videos and compresses videos with higher inter-frame redundancy.

\paragraph{Case studies.}
The two case studies on the right of Figure~\ref{fig:adaptive_behavior} visualize the same mechanism at the video level. The middle panel shows an MLVU anomaly video with a long low-\emph{pcost} interval and only late bursts, while the right panel uses a more dynamic example from NExT-QA, where multiple spikes trigger earlier I-frame refreshes.

\paragraph{From mechanism to accuracy.}
When videos remain visually stable for long periods, AdaCodec preserves much more of the original sequence within the same context budget, exposing a more complete temporal record to the MLLM. This explains the $+20.6$ MLVU anomaly-recognition gain reported in Section~\ref{sec:adaptive_gop_behavior}: the proposed \emph{pcost}-guided construction allocates longer GOPs to temporally stable videos and shorter GOPs to content with frequent scene or motion changes, matching token budget to video complexity rather than enforcing a fixed schedule.

\section{Ablation Protocol and Per-Axis Analysis}
\label{app:ablation_details}

\paragraph{Why subset-based ablations.}
Running the full AdaCodec training pipeline for every codec variant would be prohibitively expensive in both wall-clock time and GPU hours. We therefore perform ablations on a fixed subset that contains randomly sampled 1/3 of the full training corpus instead of repeating full-data training for every setting. The subset contains 1{,}301{,}438 examples, which is still large enough to support stable training and reveal the qualitative trends.

\paragraph{Protocol.}
All ablated variants use the same backbone, optimizer, learning-rate, input resolution, frame rate, and decoding protocol as the full model. We train each variant on the same reduced training split for the same number of optimization steps. This keeps the ablations focused on relative ranking among design choices rather than absolute leaderboard performance. After identifying the final operating point, we retrain that configuration on the full training data.

\begin{table}[h]
\centering
\small
\caption{Full ablation sweep over codec-design axes of AdaCodec. Each value is the mean score over the benchmarks in its category. The top row reports the subset-trained default; each subsequent row varies a single axis and shows the absolute score together with its change relative to the default in subscript. For the \emph{pcost} threshold target, $K$ denotes the number of P-frames per GOP, so the total GOP length is $K+1$ including the I-frame. The dynamic macroblock setting follows H.264 variable block size: $16{\times}16$, $16{\times}8$, $8{\times}16$, $8{\times}8$, selected per region by motion and residual complexity.}
\label{tab:codec_ablation_full}
\setlength{\tabcolsep}{4pt}
\begin{tabular}{llccc}
\toprule
Axis & Setting & Long-video$\uparrow$ & Temporal$\uparrow$ & General$\uparrow$ \\
\midrule
\multirow{2}{*}{\textbf{AdaCodec}}
  & {$N_P{=}16$, $K_{\max}{=}16$, Med.~$K{=}8$,}
                                      & \multirow{2}{*}{\textbf{60.3}} & \multirow{2}{*}{\textbf{56.2}} & \multirow{2}{*}{\textbf{74.1}} \\
  & {MB${=}16{\times}16$, adaptive GOP} &                       &                       &                       \\
\midrule
\multirow{2}{*}{P-token count $N_P$}
  & 8                                    & 60.4\dplus{0.1}  & 55.8\dminus{0.4} & 73.9\dminus{0.2} \\
  & 24                                   & 59.8\dminus{0.5} & 56.1\dminus{0.1} & 74.4\dplus{0.3}  \\
\midrule
\multirow{2}{*}{Max \#P-frames per GOP $K_{\max}$}
  & 8                                    & 59.4\dminus{0.9} & 56.2\dzero       & 74.2\dplus{0.1}  \\
  & 24                                   & 60.1\dminus{0.2} & 56.0\dminus{0.2} & 73.5\dminus{0.6} \\
\midrule
\multirow{2}{*}{\emph{pcost} threshold target}
  & Median $K=4$                         & 59.2\dminus{1.1} & 55.7\dminus{0.5} & 73.8\dminus{0.3} \\
  & Median $K=12$                        & 60.0\dminus{0.3} & 55.5\dminus{0.7} & 73.6\dminus{0.5} \\
\midrule
Macroblock size
  & Dynamic (H.264-style)          & 58.4\dminus{1.9} & 54.8\dminus{1.4} & 72.6\dminus{1.5} \\
\midrule
\multirow{2}{*}{GOP construction}
  & Fixed, $n_{P} = 8$                   & 58.2\dminus{2.1} & 56.0\dminus{0.2} & 73.7\dminus{0.4} \\
  & Fixed, $n_{P} = 16$                  & 59.7\dminus{0.6} & 55.4\dminus{0.8} & 72.3\dminus{1.8} \\
\bottomrule
\end{tabular}
\end{table}

\paragraph{Per-axis analysis.}
Table~\ref{tab:codec_ablation_full} ablates the P-token count $N_P$ (how many learned tokens represent one P-frame), the maximum number of P-frames per GOP $K_{\max}$, the \emph{pcost} threshold target used to calibrate $\gamma$, the macroblock size used for motion and residual modeling ($16{\times}16$ aligned with ViT patches versus H.264-style dynamic partitioning), and the GOP construction strategy (our adaptive \emph{pcost}-guided policy versus fixed-length baselines with $n_{P} \in \{8, 16\}$ P-frames per GOP). AdaCodec is largely insensitive to the P-token count: $N_P{=}16$ is best, but $8$ and $24$ stay within $0.5$ on every category. The largest hit appears on long-video at $N_P{=}24$ ($-0.5$), where each P-frame consumes more tokens and fewer frames fit the same budget. The same mechanism explains the chain-length sweep: $K_{\max}{=}8$ loses $0.9$ on long-video because shorter chains insert more token-heavy I-frames and shrink the usable frame count. The \emph{pcost} threshold sweep gives the strongest results at the default median $K=8$. A shorter target, median $K=4$, refreshes I-frames more often and reduces temporal coverage under the same context budget, while a longer target, median $K=12$, saves more tokens but increases the length of predictive chains, which hurts dynamic videos with larger residuals. Replacing the ViT-aligned $16{\times}16$ macroblocks with native H.264 dynamic partitioning costs $1.4$--$1.9$ across all three categories, since the multiple block sizes break the per-patch motion-vector uniformity that the P-tokenizer's patch-embedding stem relies on. Adaptive \emph{pcost}-guided GOP construction beats both fixed-length baselines on every category, with the largest gaps on long-video ($2.1$ for $n_{P}{=}8$) and general ($1.8$ for $n_{P}{=}16$); this directly matches the content-dependent GOP-length variation in Figure~\ref{fig:adaptive_behavior}, where slow-content videos (e.g., anomaly recognition) receive longer GOPs and dynamic ones receive shorter, a regime no fixed schedule can capture.

\end{document}